\documentclass[10pt,twocolumn,letterpaper]{article}

\usepackage{cvpr}              

\usepackage{graphicx}
\usepackage{amsmath}
\usepackage{amssymb}
\usepackage{booktabs}

\usepackage{times}
\usepackage{latexsym}
\usepackage{enumitem}
\usepackage{makecell}
\usepackage{lipsum}
\usepackage{multirow}
\usepackage{colortbl}
\usepackage{tabularx}
\usepackage{framed}
\usepackage{bbding}
\usepackage{lipsum}
\usepackage{longtable}
\usepackage{diagbox}
\usepackage[normalem]{ulem}
\useunder{\uline}{\ul}{}
\usepackage{CJKutf8}
\usepackage{fancyhdr}
\usepackage{float}
\usepackage{overpic}
\usepackage{bm}

\usepackage[pagebackref,breaklinks,colorlinks]{hyperref}

\usepackage[capitalize]{cleveref}
\crefname{section}{Sec.}{Secs.}
\Crefname{section}{Section}{Sections}
\Crefname{table}{Table}{Tables}
\crefname{table}{Tab.}{Tabs.}

\def\cvprPaperID{9} 
\def\confName{CVPR}
\def\confYear{2023}
\def\framework{{$\bm{A^{3}R}$}}

\begin{document}


\title{Self-Enhancement Improves Text-Image Retrieval 
in Foundation Visual-Language Models}

\author{Yuguang Yang$^{1}$\quad Yiming Wang$^{1,2}$\quad Shupeng Geng$^{1}$\quad Runqi Wang$^{1}$ \\Yimi Wang$^{1}$\quad Sheng Wu$^{1}$\quad Baochang Zhang$^{1,3,}${\thanks{Corresponding author. This paper was partially supported by the ``One Thousand Talents Plan'' innovation leading talent funding projects in Jiangxi Province, China, and National Innovation and Entrepreneurship Training Municipal Project, China.}}\\
$^1$Beihang University \quad
$^2$Shanghai Jiao Tong University  \quad
$^3$Zhongguancun Laboratory\\
{\tt\small \{guangbuaa, yiming.wang, gengshupeng, bczhang\}@buaa.edu.cn}
}

\maketitle

\begin{abstract}
The emergence of cross-modal foundation models has introduced numerous approaches grounded in text-image retrieval. 
However, on some domain-specific retrieval tasks, these models fail to focus on the key attributes required.
To address this issue, we propose a self-enhancement framework, $\bm{A^{3}R}$, based on the CLIP-ViT/G-14, one of the largest cross-modal models. 
First, we perform an \textbf{Attribute Augmentation} strategy to enrich the textual description for fine-grained representation before model learning. 
Then, we propose an \textbf{Adaption Re-ranking} method to unify the representation space of textual query and candidate images and re-rank candidate images relying on the adapted query after model learning. The proposed framework is validated to achieve a salient improvement over the baseline and other teams' solutions in the cross-modal image retrieval track of the 1st foundation model challenge without introducing any additional samples.
The code is available at \url{https://github.com/CapricornGuang/A3R}.
\end{abstract}

\section{Introduction}

Image retrieval has been widely applied in automatic public video surveillance that obtains several relevant images from vast image databases based on user queries~\cite{application}.
Traditional retrieval methods rely on attribute recognition to identify the desired images~\cite{attr-recog1, attr-recog2}, which struggle to handle the customized retrieval query and thus lack the flexibility to various user needs.
Recently, cross-modal foundation models have gained popularity for their ability to unify text and image representations~\cite{clip,vit,vitg14}, 
the large-scale pretraining data equips them with the generalization ability to handle a wide range of real-world scenes, so as to enable them with seamless translation between text and images, enhancing the flexibility and adaptability of the retrieval process.

\begin{figure}[tb]
  \centering
  \includegraphics[width=0.95\linewidth]{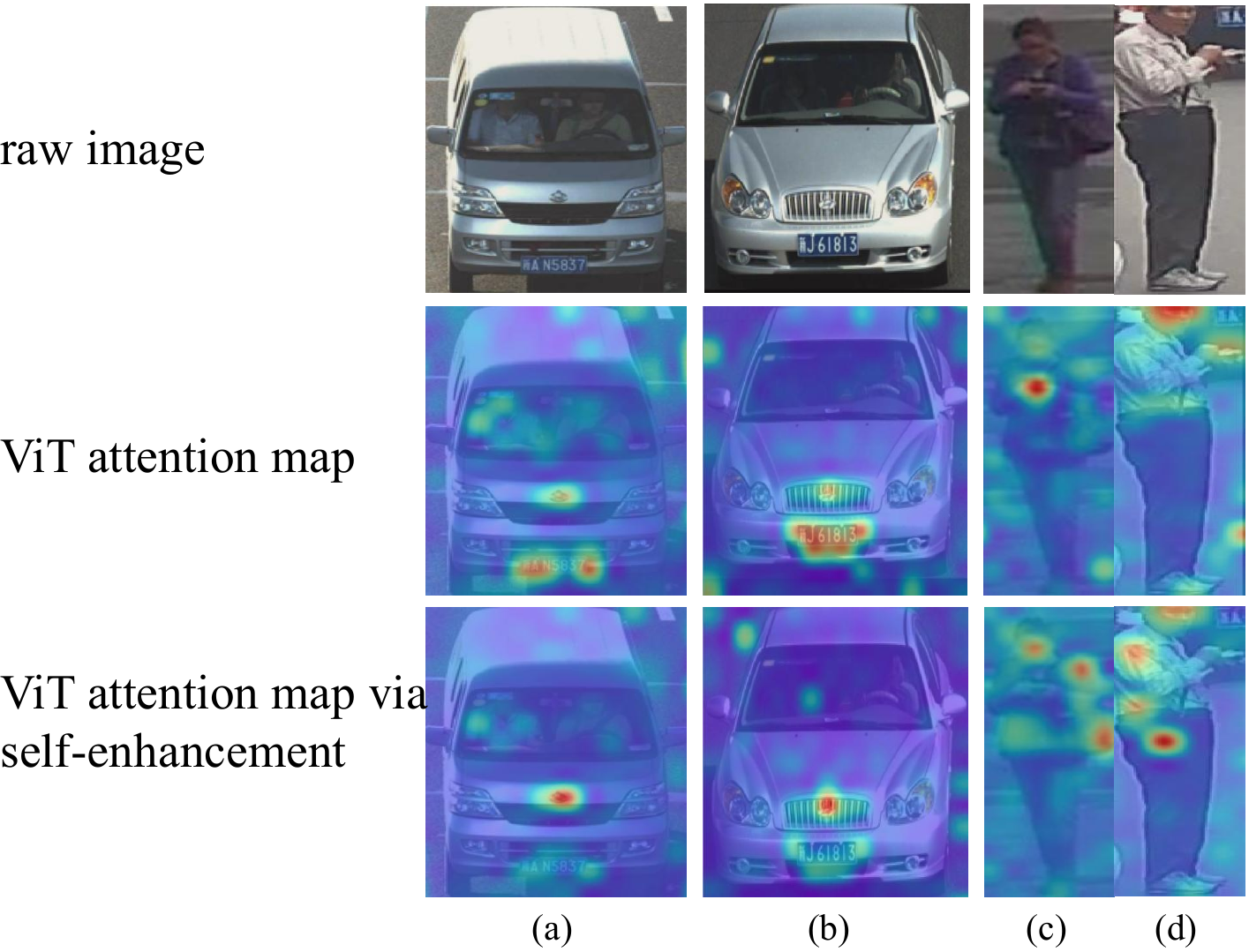}  
  \vspace{-0.1in}
  \caption{Raw images and their corresponding attention maps of CLIP-ViT/G-14 before and after self-enhancement. Before attribute augmentation, the model primarily attends to 
  sample-specific features, such as vehicle accessories, license plates, and human heads. But after attribute augmentation, the model demonstrates a notable shift towards class-specific features, specifically vehicle logos, human attire, and backpacks.
  }
  \label{img:first_page}
\end{figure}

\begin{figure*}[tb]
  \centering
  \includegraphics[width=1\linewidth]{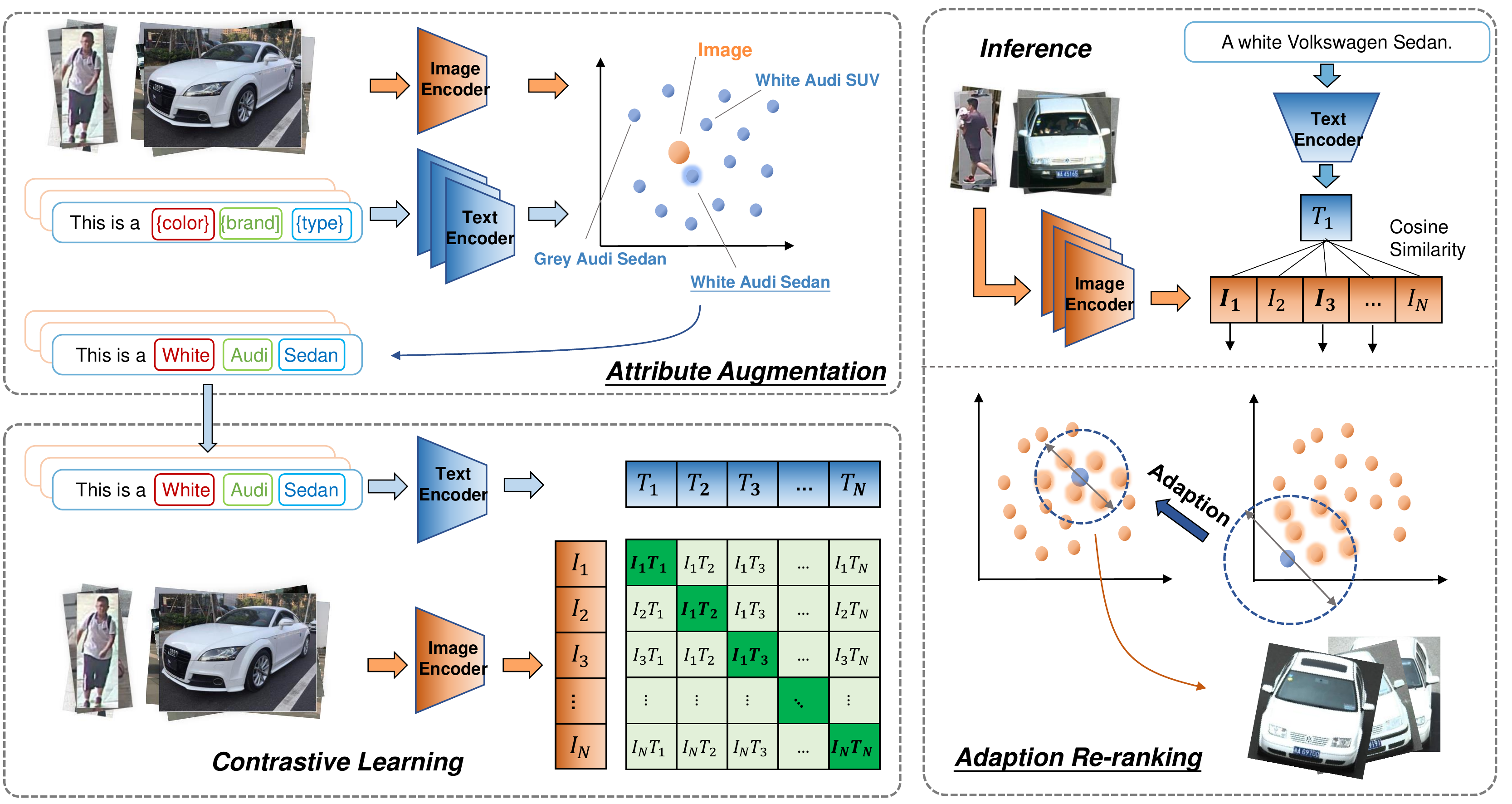}  
  \vspace{-0.25in}
  \caption{Full pipeline of $A^{3}R$ framework. \textit{Attribute Augmentation} and \textit{Adapation Re-ranking} are our self-enhancement method.}
  \label{img:pipeline}
\end{figure*}

However, as shown in Figure \ref{img:first_page}, on some class-specific retrieval tasks, the large-scale pretraining makes these foundation models focus more on the sample-specific attributes rather than the class-specific attributes required.
To address this issue, we propose a self-enhancement framework in this work, aiming to deploy the visual-language foundation model in the context of traffic retrieval to further improve the accuracy of text retrieval for images.
We perform data and retrieval augmentation before and after model learning, respectively:
(i) We recognize that the most valuable information benefiting fine-grained retrieval is the attribute description, 
{so we utilize the rich prior knowledge of foundation models to perform zero-shot \textbf{\textit{Attribute Augmentation}} to augment the textual description with various attributes};
(ii) We note that similarities between candidate images often overshadow the similarities between images and textual queries in cross-modal retrieval,
so we propose the \textit{\textbf{Adaption Re-ranking}} method to extract query-specific information from candidate images and instead leverage the original textual query as a similarity constraint.
We first align the representation space between the re-ranking query and candidate images, then perform cross-modal re-ranking to retrieve the final expected images. By combining these two strategies, We name our framework \framework{}.
Experiments show that our $A^3R$ framework enables models to achieve salient improvements over the pure fine-tuning paradigm.

\section{$A^3R$: Self-Enhancement Framework}

We resort to CLIP-ViT/G-14 \cite{vitg14} for our backbone (§\ref{sec:backbone}), which achieves dramatic cross-modal performances in many tasks. However, it performs poorly in domain-specific tasks like cross-modal retrieval.
Therefore, we propose a zero-shot self-enhancement framework based on attribute augmentation (§\ref{sec:aa}) and adaption re-ranking (§\ref{sec:re}) to improve the capability of fine-tuned visual-language models.
Figure \ref{img:pipeline} illustrates our framework.

\subsection{Architecture and Formulation}
\label{sec:backbone}
We formulate each sample as a package of image, text, and attribute $(I, T, A)$, where attributions can be viewed as the subset of texts that present the most distinguishing information, such as ``\textit{color}'', etc.
First, $I$ and $T$ are input to image and text encoders, respectively, which are both transformer-based modules generating image and text embeddings $w^{I}$ and $w^{T}$.
Then, contrastive learning is employed in the following training.
Specifically, assume that $N$ encoded samples $\{(w^{I}_i, w^{T}_i)\}_{i=1}^N$ are processed in parallel. Let image embedding matrix be $M_I = \{w^{I}_i\}_{i=1}^N$ and text embedding matrix be $M_T = \{w^{T}_i\}_{i=1}^N$, then the text-image similarity matrix $F$ are formulated as
\begin{equation}
    F = (M_I) (M_T)' \in \mathbb{R}^{N \times N}.
\end{equation}
The fine-tuning target is to maximize $N$ matched pairs' similarity and minimize that of $N^2-N$ mismatched pairs.

\subsection{Attribute Augmentation}
\label{sec:aa}
We note that the most valuable information in the texts is the attributes. 
However, they are often missing in the labels of vehicle datasets. Therefore, we perform zero-shot data augmentations aimed at filling in the missing attribute elements in the samples.
We focus on three attribute elements: \textit{color}, \textit{brand}, and \textit{type}. For the missing attribute element, we exhaust all possible cases of this attribute element and concatenate them after the existing text $T$, respectively, and then pass all concatenated text and the corresponding only image into the CLIP-ViT/G-14 model to obtain embeddings of the image and all texts.
We select the text with the highest image-text cosine similarity as the augmented text $T$.

\subsection{Adaption Re-ranking}
\label{sec:re}
Re-ranking is vital for improving retrieval performance \cite{retrieval-survey}. This process typically involves two key steps: re-sorting query embedding and similarity refinement.
However, in cross-modal retrieval, the re-sorting query can not be directly taken from the textual query due to the contrastive-based language-image alignment strategy. This strategy aligns different modalities without emphasizing the need for shared representation spaces, leading to the similarities between candidate images often overshadowing the similarities between images and textual queries.

To address this, we propose Adaption Re-ranking, a plug-and-play non-parameter modal adapter utilizing singular vector decomposition (SVD) to extract query-specific information from candidate images considering the text-image similarity. Specifically, given the encoded textual query embedding $M_q = \{w^{T}_0\}$ and $M$ candidate encoded image embeddings $M_c = \{w^{I}_i\}_{i=1}^{M}$ processed in parallel, we first compute the query-candidate similarity as follows:
\begin{equation}
S = (M_c) (M_q)' \in \mathbb{R}^{M \times 1}.
\end{equation}
To simplify computations, we broadcast the query-candidate similarity metric $S$ into $\widetilde{S} = \{S\}_{i=1}^{D}\in \mathbb{R}^{M\times D}$. SVD is then performed on $M_c$ under the constraint of similarity $S$, yielding: 
\begin{equation}
M_s = M_c \odot \widetilde{S} = U \Sigma V^{T},
\end{equation}
where $U \in \mathbb{R}^{M\times M}$, $V \in \mathbb{R}^{ D\times D}$ are unitary matrixes, and $\Sigma \in \mathbb{R}^{M\times D}$ is a diagonal matrix whose diagonal entries correspond to the singular values of $M_s$ sorted in descending order. 
By utilizing the column vector $U_{1} \in \mathbb{R}^{M\times 1}$ of $U$ \textit{w.r.t.} the largest singular value, the principal vector of the encoded candidate images $M_{q}^{\ast}\in \mathbb{R}^{1\times D}$ can be represented as $M_{q}^{\ast} = U_{1}' M_c$, which is used as the re-ranking query. 

Subsequently, we incorporate the k-reciprocal encoding algorithm\cite{krnn} to further refine the re-ranking results. The k-reciprocal encoding algorithm is a well-established  re-ranking technique. It aims to calculate a new similarity measure between a query image and a candidate image, based on their k-reciprocal nearest neighbors:
\begin{equation}
M_{c}^{\ast} = k\text{-reciprocal}(M_{q}^{\ast}, M_{c}).
\end{equation}

\section{Experiments}
\subsection{Setup}

\paragraph{Dataset.}
We use the open-source \textbf{PA100k} pedestrian dataset\cite{pa100k} and the \textbf{BIT-Vehicle} vehicle dataset\cite{bitvehicle}.
Unlike previous approaches relying on one-hot encoded attribute labels, these datasets incorporate attribute-level natural language text annotations corresponding to the images.
The pedestrian dataset is constructed by human images of 21 unique attributes concerning sex, dress, and walking position, while the vehicle dataset contains auto images of 11 colors, 6 vehicle types, and 65 vehicle brands.
We note that the vehicle images in the test set were obtained through web scraping, resulting in significant variations in data distribution, while that of pedestrian images is relatively consistent.

\paragraph{Metric.}
The evaluation metric used is the mean Average Precision ($mAP@K$). Here, $K$ indicates that the top$K$ retrieval results are used in the evaluation. The calculation of ${mAP@K}$ is as follows:
\begin{equation}
    {mAP@K} = \frac{1}{m} *\sum_{i=1}^{K} p(i) * \Delta r(i),
\end{equation}
where ${m}$ is the total query times in the evaluation set, ${p(i)}$ and $r(i)$ denotes the precision and recall of the top${i}$ retrieval results, respectively, ${\Delta r(i)} = r(i) - r(i-1)$, and ${r(0)=0}$.

\paragraph{Implementation.}
We adopt the 2B-parameter CLIP-ViT/G-14\cite{vitg14} as our backbone.
To address the variations in image proportions between vehicles and pedestrians, we apply zero-padding and resize all images to a size of 224. The training process utilizes the Adam optimizer with a batch size of 25.
Given the significant distribution disparities between the training and test sets, the performance improvement on the validation set can not be reflected on the test set directly.
Therefore, we employ a small learning rate of 4e-7 and perform only 5 epochs of fine-tuning.
The training process utilizes one A100 GPU, while the inference process is executed on one NVIDIA RTX 3090 GPU.

\subsection{Main Results}
The leaderboard ranking is determined based on the calculation of ${mAP@10}$. Our competition performance is shown in Table \ref{tab:team-rank}. Our proposed method ranks 6th on the leaderboard with a score of 0.75027 without introducing any additional datasets. 
These results indicate that $A^3R$ can capture both textual and visual information, enabling accurate retrieval in real-world scenarios.

\begin{table}[htb]
  \centering
  \resizebox{0.93\columnwidth}{!}{
  \begin{tabular}{ccc|ccc}
    \toprule
    Rank & Team Name & Score & Rank & Team Name & Score\\
    \midrule
    1 & MiniModel & 0.82382 & \textbf{6} & \textbf{IPCL(ours)} & \textbf{0.75027}\\
    2 & njust & 0.82223 & 7 & VIPS & 0.74710\\
    3 & DiamondH & 0.81990 & 8 & 432 & 0.73654\\
    4 & CASHIPS & 0.76865 & 9 & BBH$\sim$ & 0.72753\\
    5 & HZHv2 & 0.76268 & 10 & mARapper & 0.72697\\

    \bottomrule
  \end{tabular}}
  \caption{Leaderboard A of the Cross-modal Track in the Foundation Model Challenge.}
  \label{tab:team-rank}
\end{table}

\subsection{Analysis}
\begin{figure}[tb]
  \centering
  \includegraphics[width=1\linewidth]{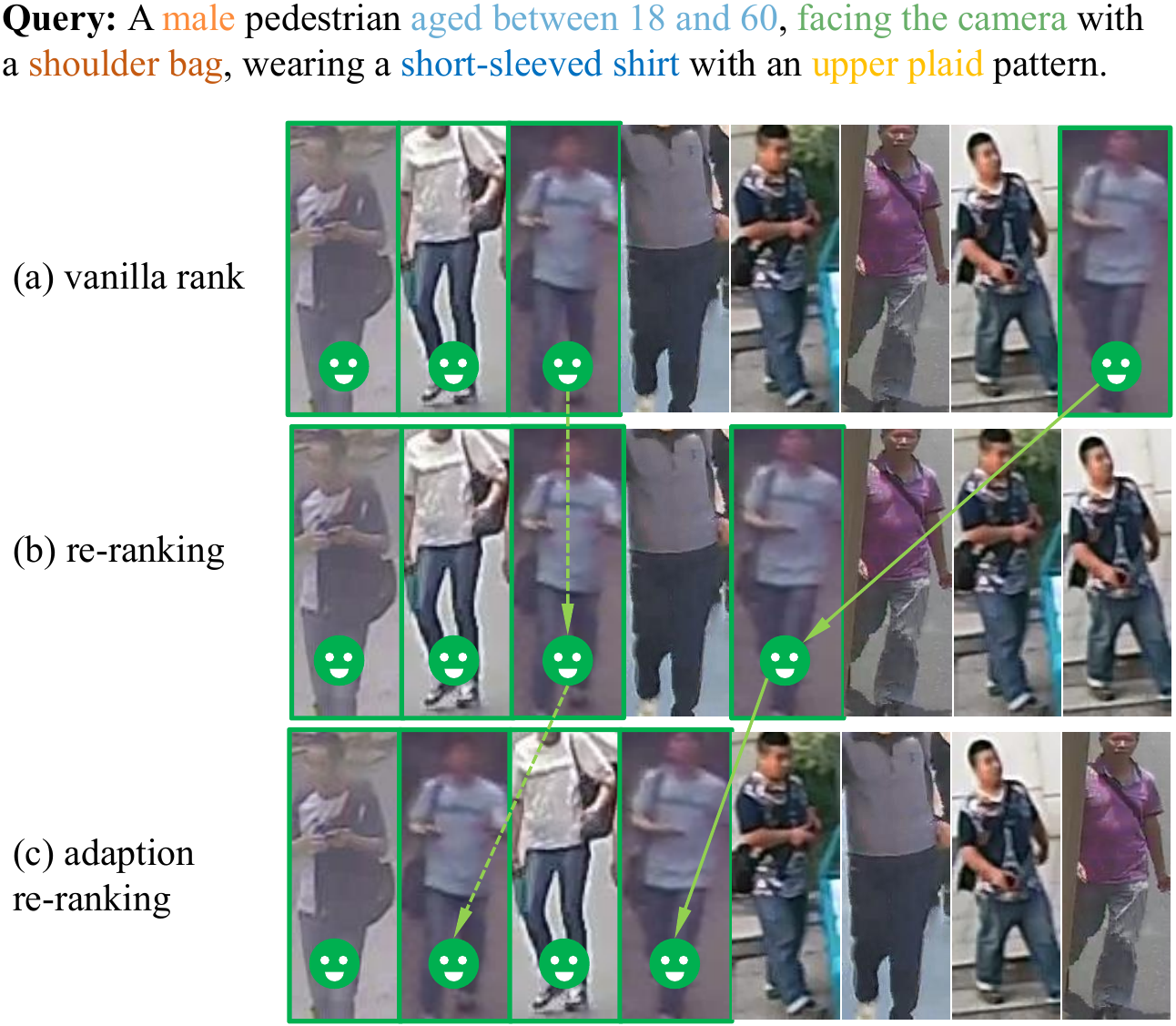}  
  \vspace{-0.25in}
  \caption{
  Visualization of the re-ranking algorithm. 
  \includegraphics[height=1em]{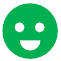} indicates the query-matched objects.
  }
  \label{img:reranking}
\end{figure}

\paragraph{Re-ranking.} 
Figure \ref{img:reranking} illustrates the effect of the re-ranking algorithm. 
As shown in sub-figure (a), the vanilla searching results have been already accurate to some extent, but some relevant objects are not ranked at the top.
Comparing sub-figure (b) and (c), we observe that the proposed adaptation re-ranking strategy effectively promotes lower-ranked target samples to higher positions, as shown by the solid lines in the figure.
Furthermore, our method can further improve the retrieval performance by fine-tuning the already well-ranked correct retrievals, as shown by the dashed lines in the figure.

\begin{figure}[tb]
  \centering
  \includegraphics[width=1\linewidth]{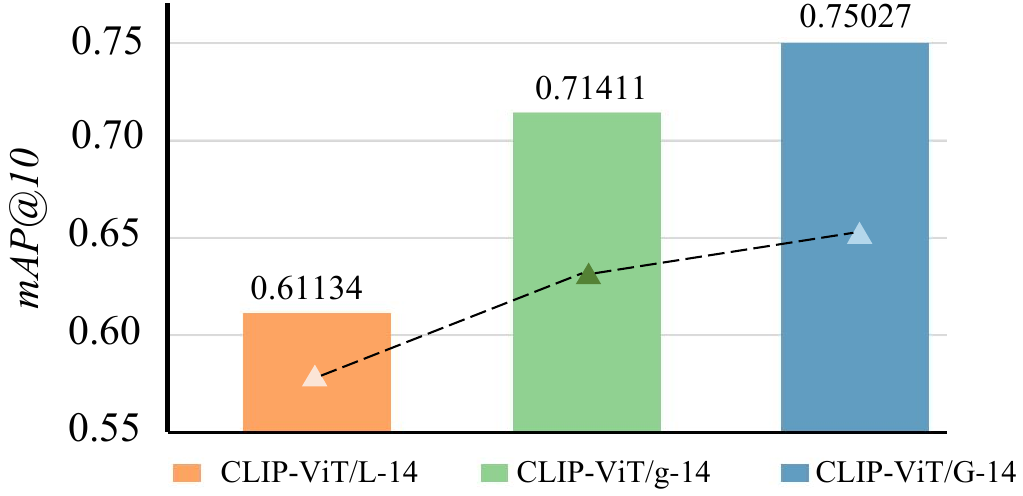}
  \vspace{-0.25in}
  \caption{Comparison of the $mAP@10$ metric of backbones of different parameter sizes.}
  \label{img:parameter}
\end{figure}

\paragraph{Model Parameter.} 
Figure \ref{img:parameter} illustrates the performance of CLIP based on ViT/L-14, ViT/g-14, and ViT/G-14. ViT/g-14 exhibits approximately five times higher computational complexity than ViT/L-14, while ViT/G-14 shows a seven-fold increase. 
Analyzing the performance increase at each expansion fold, we observe a slight decline in performance improvement from the second expansion compared to the first expansion (2\% $\rightarrow$ 1.8\%).
Considering the challenge associated with ultra-large models, this evidence indicates that current visual foundation models have room for improvement before reaching their performance limits.

\section{Conclusion}
This paper explores a new avenue to improve the performance of foundation models with their inner knowledge, so-called ``self-enhancement".
The proposed $A^3R$ framework performs zero-shot attribute augmentation to augment the downstream dataset before model learning while harnessing the internal relationship between the text and image representation space to conduct cross-modal re-ranking to retrieve the final expected images. This self-enhancement framework can be transferred to any cross-modal retrieval scenario with good domain generality.

{\small
\bibliographystyle{ieee_fullname}
\bibliography{egbib}

\begin{thebibliography}{10}\itemsep=-1pt

\bibitem{vit}
Alexey Dosovitskiy, Lucas Beyer, Alexander Kolesnikov, and et. al.
\newblock An image is worth 16x16 words: Transformers for image recognition at
  scale.
\newblock In {\em International Conference on Learning Representations}, 2021.

\bibitem{retrieval-survey}
Mayuri~D. Joshi, Revati~M. Deshmukh, Kalashree~N. Hemke, and et. al.
\newblock Image retrieval and re-ranking techniques-a survey.
\newblock {\em SIPIJ}, 5(2), 2014.

\bibitem{application}
Rajiv Kapoor, Deepak Sharma, and Tarun Gulati.
\newblock State of the art content based image retrieval techniques using deep
  learning: a survey.
\newblock {\em Multimedia Tools and Applications}, 80(19):29561--29583, 2021.

\bibitem{pa100k}
Pengze Liu, Xihui Liu, Junjie Yan, and Jing Shao.
\newblock Localization guided learning for pedestrian attribute recognition.
\newblock In {\em British Machine Vision Conference 2018}, page 142. {BMVA}
  Press, 2018.

\bibitem{krnn}
Danfeng Qin, Stephan Gammeter, Lukas Bossard, and et. al.
\newblock Hello neighbor: Accurate object retrieval with k-reciprocal nearest
  neighbors.
\newblock In {\em The 24th {IEEE} Conference on Computer Vision and Pattern
  Recognition, {CVPR} 2011}, pages 777--784. {IEEE} Computer Society, 2011.

\bibitem{attr-recog2}
Rodolfo Quispe, Cuiling Lan, Wenjun Zeng, and et. al.
\newblock Attributenet: Attribute enhanced vehicle re-identification.
\newblock {\em Neurocomputing}, 465:84--92, 2021.

\bibitem{clip}
Alec Radford, Jong~Wook Kim, Chris Hallacy, and et. al.
\newblock Learning transferable visual models from natural language
  supervision.
\newblock In Marina Meila and Tong Zhang, editors, {\em Proceedings of the
  International Conference on Machine Learning}, volume 139 of {\em Proceedings
  of Machine Learning Research}, pages 8748--8763. {PMLR}, 2021.

\bibitem{bitvehicle}
Jun Sang, Zhongyuan Wu, Pei lin Guo, and et. al.
\newblock An improved yolov2 for vehicle detection.
\newblock {\em Sensors}, 18, 2018.

\bibitem{attr-recog1}
Andreas Specker, Mickael Cormier, and J{\"u}rgen Beyerer.
\newblock Upar: Unified pedestrian attribute recognition and person retrieval.
\newblock In {\em WACV}, pages 981--990, 2023.

\bibitem{vitg14}
Xiaohua Zhai, Alexander Kolesnikov, Neil Houlsby, and et. al.
\newblock Scaling vision transformers.
\newblock In {\em CVPR}, pages 12104--12113, 2022.

\end{thebibliography}
}

\end{document}